\documentclass[preprint, review,3p,times, 10pt, authoryear]{elsarticle}

\usepackage{cite}
\usepackage{amsmath,amssymb,amsfonts}
\usepackage{algorithmic}
\usepackage{graphicx}
\usepackage{textcomp}
\usepackage{comment}
\usepackage{footnote}
\usepackage{comment}
\usepackage{threeparttable}
\usepackage{romannum}
\usepackage{longtable}
\usepackage{lscape}
\usepackage{soul}
\usepackage{array, makecell}
\usepackage{tikz}
\usepackage{pgfplots}
\usepackage{pgf-pie}
\usepackage{longtable}

\usepackage{amssymb}
\usepackage{utfsym}
\newcommand{\crossmark}{\scalebox{0.75}{\usym{2613}}}

\usepackage{caption}
\usepackage{booktabs,tabularx}
\usepackage{multicol, multirow}
\newcommand{\figref}[1]{\figurename~\ref{#1}}
\newcommand{\tabref}[1]{\tablename~\ref{#1}}
\usepackage{multicol}
\usepackage[hyphens]{url}
\usepackage{hyperref}
\hypersetup{colorlinks=true,breaklinks=true}

\journal{arxiv}

\begin{document}

\begin{frontmatter}
\title{MMTF-DES: A Fusion of Multimodal Transformer Models for Desire, Emotion, and Sentiment Analysis of Social Media Data}

\author[1]{Abdul Aziz}\ead{aziz.abdul.cu@gmail.com}
\author[1]{Nihad Karim Chowdhury\corref{correspondingauthor}}\ead{nihad@cu.ac.bd}
\cortext[correspondingauthor]{Corresponding author}
\author[2]{Muhammad Ashad Kabir}
\ead{akabir@csu.edu.au}%
\author[1]{Abu Nowshed Chy}\ead{nowshed@cu.ac.bd}
\author[3]{Md. Jawad Siddique}
\ead{mdjawad.siddique@siu.edu}%
\affiliation[1]{organization={Department of Computer Science and Engineering, University of Chittagong}, city={Chattogram}, postcode={4331},country={Bangladesh}}
\affiliation[2]{organization={School of Computing, Mathematics and Engineering, Charles Sturt University}, city={Bathurst}, state={NSW}, postcode={2795}, country={Australia}}
\affiliation[3]{organization={Department of Computer Science, Southern Illinois University}, city={Carbondale}, state={Illinois}, postcode={62901},  country={USA}}

\begin{abstract}
Desire is a set of human aspirations and wishes that comprise verbal and cognitive aspects that drive human feelings and behaviors, distinguishing humans from other animals. Understanding human desire has the potential to be one of the most fascinating and challenging research domains. It is tightly coupled with sentiment analysis and emotion recognition tasks. It is beneficial for increasing human-computer interactions, recognizing human emotional intelligence, understanding interpersonal relationships, and making decisions. However, understanding human desire is challenging and under-explored because ways of eliciting desire might be different among humans. The task gets more difficult due to the diverse cultures, countries, and languages. Prior studies overlooked the use of image-text pairwise feature representation, which is crucial for the task of human desire understanding. In this research, we have proposed a unified multimodal transformer-based framework with image-text pair settings to identify human desire, sentiment, and emotion. The core of our proposed method lies in the encoder module, which is built using two state-of-the-art multimodal transformer models. These models allow us to extract diverse features. To effectively extract visual and contextualized embedding features from social media image and text pairs, we conducted joint fine-tuning of two pre-trained multimodal transformer models: Vision-and-Language Transformer (ViLT) and Vision-and-Augmented-Language Transformer (VAuLT). Subsequently, we use an early fusion strategy on these embedding features to obtain combined diverse feature representations of the image-text pair. This consolidation incorporates diverse information about this task, enabling us to robustly perceive the context and image pair from multiple perspectives. Moreover, we leverage a multi-sample dropout mechanism to enhance the generalization ability and expedite the training process of our proposed method. To evaluate our proposed approach, we used the multimodal dataset MSED for the human desire understanding task. Through our experimental evaluation, we demonstrate that our method excels in capturing both visual and contextual information, resulting in superior performance compared to other state-of-the-art techniques. Specifically, our method outperforms existing approaches by 3\% for sentiment analysis, 2.2\% for emotion analysis, and approximately 1\% for desire analysis.
\end{abstract}

\end{frontmatter}

\section{Introduction}
\label{sec:introduction}
Social media platforms such as Twitter, Reddit, Facebook, and Instagram, have increasingly become the most widely used means of information sharing due to their practical features and real-time behavior~\citep{CAMACHO202088}. People typically present their thoughts, opinions, breaking news, and concepts using various information modalities such as text, visual, and audio~\citep{AFYOUNI2022279}. The reliability of social media makes it an engaging source of information for researchers and business organizations to discover knowledge~\citep{zhang2022big}. Over the last decade, text-based sentiment analysis~\citep{xu2019sentiment,alamoodi2021sentiment,rani2022efficient} and emotion recognition~\citep{sailunaz2019emotion,estrada2020opinion,khoshnam2022dual} from social media have advanced quickly and drawn a lot of interest from both academia and industry. Multimodal sentiment analysis~\citep{morency2011towards,soleymani2017survey,zhu2022multimodal} and emotion recognition~\citep{poria2016convolutional,ranganathan2016multimodal,middya2022deep} gained greater interest because of their multimodal nature which makes them more understandable. However, understanding human sentiments and emotions across several modalities is challenging because of the complexity of human feelings and expressions.

Desire, a fundamental aspect of human nature, encompasses a strong feeling of craving or longing for something~\citep{dong2010human}. It distinguishes humans from other animals, as they are uniquely driven by the desire to acquire or possess something, and can exhibit an unquenchable thirst. This desire motivates individuals to act and work toward their goals~\citep{wilt2015affect}. It is an innate sense that permeates all human beings and holds power to shape and influence various aspects of human behavior.

Desires often drive human sentiment and emotions, whether it be the desire for inspiration, anticipation, letdown, jealousy, or obsession. These emotions significantly shape our practical life experiences, influencing our thoughts, actions, and decisions. There is a close connection between human desire, sentiment, and emotion. Desire implicitly governs sentiment and emotion, while sentiment and emotion, in turn, can be influenced by desire. Together, desire, sentiment, and emotion form interconnected and essential components of the human experience, driving our actions and decisions.

\begin{figure*}[htbp]
\centering
\includegraphics[width=.9\linewidth]{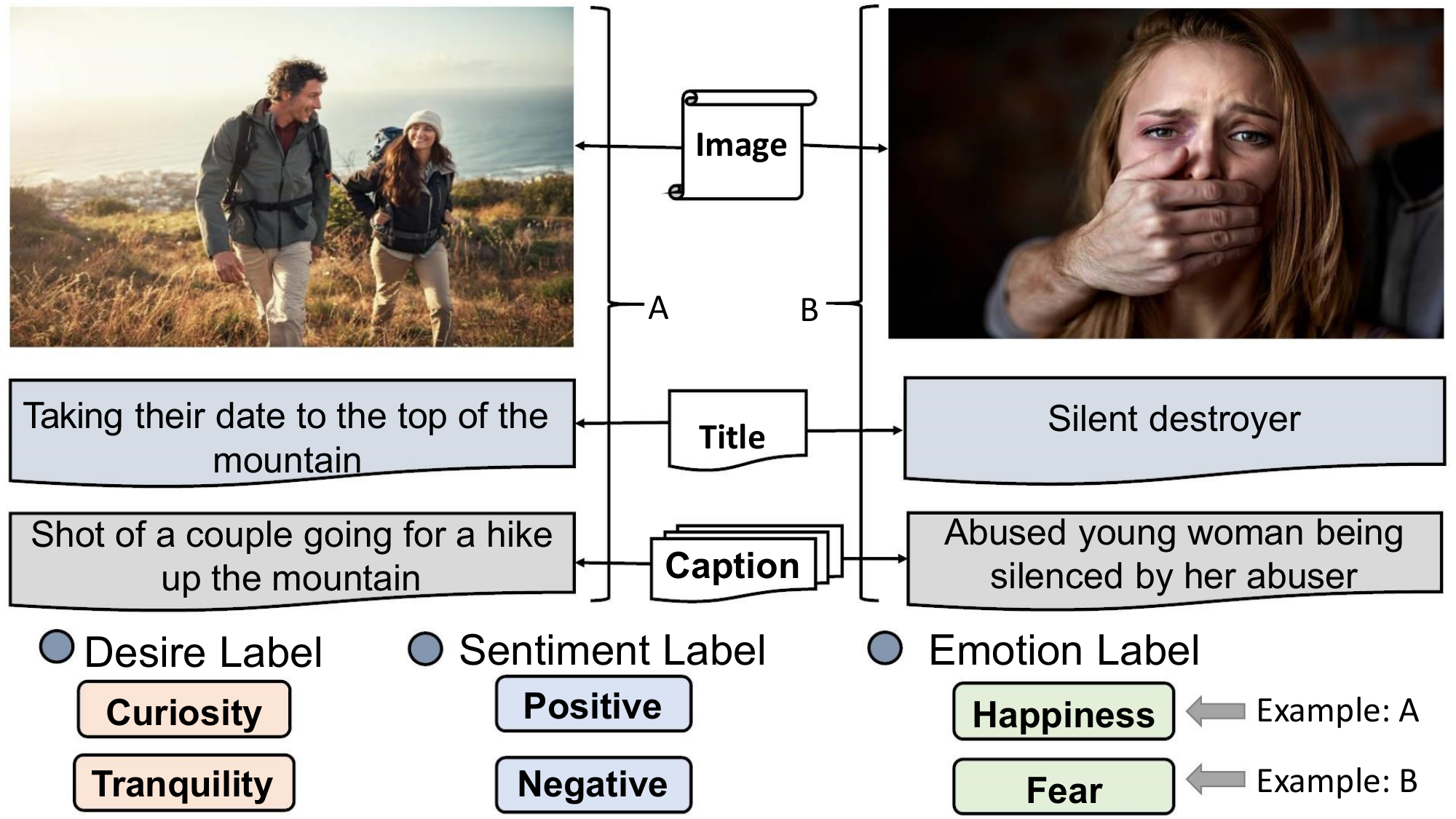}
\caption{Example of the multimodal human desire understanding task.}
\label{fig:Example}
\end{figure*}

Multimodal desire understanding is an emerging task that plays an important role in recognizing human emotional intelligence and personalized and effective human-computer interaction, leading to improved customer satisfaction and experience in e-commerce. However, the multifaceted nature of human desire makes it difficult to understand it across several modalities.  \citet{jia2022beyond} introduced the first benchmark dataset MSED focusing on the multimodal human desire understanding task. They designed the task to consist of three subtasks - desire analysis, sentiment analysis, and emotion analysis. The examples of the human desire understanding task in~\figref{fig:Example} indicate that three classification labels are required for a given image, title, and caption. A couple climbing a mountain is depicted by the image and text combination in this figure from example A. To satisfy their curiosity, they climb to the top of the mountain. According to this example, it can be observed that the sentiment and the emotion (happiness) are driven by the desire (curiosity) to achieve their goals (to reach the top of the mountain). In example B, the image-text pair illustrates how a young woman was silenced by her abuser. Since the scenario shows a negative sentiment, the woman is feeling fear. As a result, her desire is for tranquility from her attacker.

In recent years, multimodal sentiment analysis~\citep{birjali2021comprehensive,chan2023state} and emotion recognition~\citep{zhang2020emotion,ezzameli2023emotion} have gained significant attention due to the growing use of multimedia data in a variety of applications, including marketing, customer service, dialogue analysis and generation, and human-computer interaction. Most sentiment and emotion analysis studies~\citep{li2020exploring,tu2022sentiment,zhang2022deep,zhang2023multitask} have used the deep learning-based approach. However, they do not focus on the multimodal human desire understanding task. We proposed a multimodal transformer fusion-based approach to fill the multimodal human desire understanding research gaps.
The key contributions of this paper  are:
\begin{itemize}
\item We propose a unified architecture of multimodal transformer models, called MMTF-DES (Multimodal Transformers Fusion for Desire, Emotion, and Sentiment analysis) for the multimodal human desire understanding task. In this proposed architecture, we jointly fine-tune two state-of-the-art pre-trained multimodal transformer models, Vison-and-Language Transformer (ViLT)~\citep{kim2021vilt} and Vision-and-Augmented-Language Transformer (VAuLT)~\citep{chochlakis2022vault}, creating a unified architecture.
\item We adopt a training strategy known as multi-sample dropout to enhance the generalizability and training efficiency of the base model. This strategy helps improve the overall performance of the MMTF-DES model.
\item We conduct a comparative analysis among other state-of-the-art models to evaluate the effectiveness of our proposed model. Also, we present a performance evaluation of the early and late fusion strategies for each of the three tasks. To ensure precise results, we propose a component-based analysis scheme that aids in determining the final task-specific labels.
\item We introduce a new task as an extension of human desire understanding research on the MSED dataset. This task aims to deepen our understanding of human desire and facilitate more effective analysis in this domain.


\end{itemize}
The rest of the paper is organized as follows: Section~\ref{sec:relatedwork} provides an overview of prior research that motivates us to work in this domain. We raise some research questions that have driven our work on this task in Section~\ref{sec:rqs}. In Section~\ref{sec:proposedMethod}, we introduce our proposed fusion of a multimodal transformer-based approach for the multimodal human desire understanding task.  We discuss experimental settings in Section~\ref{sec:experimentalsettings}. Section~\ref{sec:experimentsAndEvaluation} includes the detailed experiments and evaluation and a comparative performance analysis of the related methods. We provide some precious insights for this task in Section~\ref{sec:discussion}. We conclude our work and discuss future directions in Section~\ref{sec:conclusionAndFutureDirection}.

\section{Related work}
\label{sec:relatedwork}
Multimodal sentiment analysis and emotion recognition pertain to analyzing a person's or group's emotions and sentiments using modalities, including text, speech, facial expressions, and body language. Because of its potential uses in several industries, including healthcare, education, and social media monitoring, this discipline has recently attracted increasing interest~\citep{arano2021multimodal}.

In earlier stages, textual sentiment analysis~\citep{zhou2013sentiment,kiritchenko2014sentiment} and emotion recognition~\citep{lee2012towards,shaheen2014emotion,kratzwald2018deep} were very popular. For example, \citet{zhou2013sentiment} and \citet{ gamallo2014citius} used lexicon-based features and statistical machine learning classifiers, including support vector machine (SVM) and Na{\"\i}ve-Bayes, for the sentiment analysis of English tweets. \citet{oberlander2018analysis} introduced a dataset and provided insight into different models for emotion classification on text. A recent study used the BERT model for sentiment analysis on tweet text of coronavirus data~\citep{singh2021sentiment}. 
Recently, multimodal sentiment analysis and emotion recognition have gained the attention of researchers because of their multimodal nature. \citet{truong2019vistanet} explored the use of deep learning models for multimodal sentiment analysis. They proposed a visual aspect attention network (VistaNet) that could effectively incorporate multiple modalities, including text and image, for sentiment analysis. \citet{mai2022hybrid} proposed a hybrid contrastive learning framework of tri-modal representation, including text, image, and speech, for multimodal sentiment analysis. To reduce the modality gap, they conducted intra-inter-modal contrastive learning and semi-contrastive learning so that the model can explore cross-modal interactions. \citet{yang2022multimodal} proposed a multimodal translation for sentiment analysis (MTSA) leveraging text, visual, and audio modalities. To improve the quality of visual and audio features, they translate them into text features using BERT. \citet{peng2023fine} represented a modal label-based multi-stage method for sentiment analysis. They consider different modalities including unimodal, bimodal, and multimodal, as independent tasks for multi-stage training.

For a long time, emotion recognition has been a topic of ongoing research. With the growth of image-text data owing to the emergence of social media platforms, multimodal emotion recognition on social media data has recently gained popularity. \citet{soleymani2011multimodal} applied a modality fusion strategy, using a support vector machine as the classifier. \citet{poria2016convolutional} used deep convolutional neural networks to extract features of visual and textual modalities. \citet{ranganathan2016multimodal} tried convolutional deep belief networks (CDBNs) that generate robust multimodal features of emotion expressions. \citet{nemati2019hybrid} introduced a hybrid approach to latent space data fusion in which the textual modality is combined with the auditory and visual modalities using a Dempster-Shafer (DS) theory-based evidentiary fusion methodology, and their projected characteristics are combined using a latent space linear map. Recently, \citet{zhang2022multimodal} used the manifold learning method to extract low-dimensional embedding features from high-dimensional features. Then, these features are fed into the deep convolutional neural network for emotion recognition. To learn a joint multimodal representation for emotion recognition, \citet{le2023multi} fuse multiple modalities including video frames, audio signals, and text subtitles through a unified transformer architecture.

Desire is inherently related to human sentiment and emotion, and understanding these relationships can provide insights into human behavior and decision-making, which drive the sentiments and emotions of humans. \citet{jia2022beyond} argue that there exists a significant interrelation between human desire, sentiment, and emotion, wherein desire holds a surreptitious hegemony over sentiment and emotion, while sentiment and emotion are subject to the influence of desire. However, human desire understanding is an under-explored task. Human desire involves both linguistic expression and visual expression. \citet{lim2012desire} developed a system for desire understanding based on four types of emotions - speed, intensity, regularity, and extent from human emotional voices. To investigate the variations and similarities in the association between sexual desire and love, \citet{cacioppo2012common} introduced a multilevel kernel density fMRI analytic technique. \citet{yavuz2019data} proposed a data mining approach for desire and intention using neural and Bayesian networks. However, those works have some limitations in representing the human desire understanding task effectively. To break the research gap in the understanding and detection of human desire, \citet{jia2022beyond} proposed the first multimodal dataset MSED for human desire understanding. They represent three tasks - desire analysis, sentiment analysis, and emotion analysis in this dataset which contains textual and visual modalities. This facilitates a new window in the human desire understanding research. They also provide various strong baselines based on different combinations of feature representations using various visual and textual encoders. The unified architecture of the BERT and ResNet model achieves top performance for them.

However, most of the systems struggled to achieve intra-and-inter relationships between modalities as they fused different modalities externally. These systems do not properly capture the semantic orientation of the textual information and visual representation to estimate human desire understanding. To learn the intra-and-inter relationship between modalities, pairwise training of different data, including visual and textual, is crucial for the human desire understanding task. Transformer-based systems distill contextual and visual information effectively. However, the fusion of a pre-trained vision and text transformer-based multimodal model without the pairwise training of the text inputs and joint training of image-text input struggles to distill intra-relationships between modalities. To mitigate this issue, we use multimodal transformer models with the pairwise training of text and images, which enables us to achieve the intra-relationship between image and text. We fuse two multimodal transformer models to extract diverse integrated visual-contextual features representation, which helps us achieve the inter-relationship of two modalities. Moreover, we use a multi-sample dropout training strategy to improve the 
generalizability of our proposed model.

\section{Research questions}
\label{sec:rqs}
To contribute to the state-of-the-art in the newly introduced multimodal human desire understanding task, we propose a unified multimodal neural network architecture to understand the human desires from the given image and the associated text. Also, our work involves identifying which integrated visual-contextual representation and techniques perform better. Therefore, we articulate some research questions (RQs) related to the task of human desire understanding.

\begin{itemize}
\setlength{\itemsep}{1pt}
\item \textbf{RQ1:} Can a unified multimodal neural network model capture better visual-contextual features than a single model for image-text data in the human desire understanding task?\\
\textemdash~We use two SOTA multimodal transformer models which give diverse visual-contextual representations and improve the generalizability of our proposed method. The corresponding details are available in Section~\ref{sec:rq1}.

\item \textbf{RQ2:} What is the effect of different fusion techniques on the MSED benchmark dataset?\\
\textemdash~We fuse two model feature vectors to get the unified architecture of our proposed method for the human desire understanding task. We answer this question in Section~\ref{sec:rq2}.

\item \textbf{RQ3: } How much does our proposed approach improve performance compared to the SOTA human desire understanding methods?\\
\textemdash~We evaluate the performance of our proposed system with SOTA human desire understanding methods in the experiments and evaluation section. The discussion is available in Section~\ref{sec:rq3}.

\item \textbf{RQ4:} Can different training strategies improve the base model performance for the human desire understanding task?\\
\textemdash~To enhance the performance of the base method, we use a training strategy. The significance of this training strategy is given in the discussion section. The analysis is available in Section~\ref{sec:rq4}.

\item  \textbf{RQ5:} How can the task be further extended to improve its effectiveness and applicability in the desire understanding task?\\
\textemdash~To further extend the desire understanding task, we proposed a new binary desire analysis task. The details of this task description and performance are analyzed in \ref{sec:rq5}.
\end{itemize}
In addition, we present a summarized analysis of these five RQs in Section~\ref{sec:researchfindings} to analyze the potency of our proposed method.

\begin{figure*}[t]
\centering
\includegraphics[width=.9\linewidth]{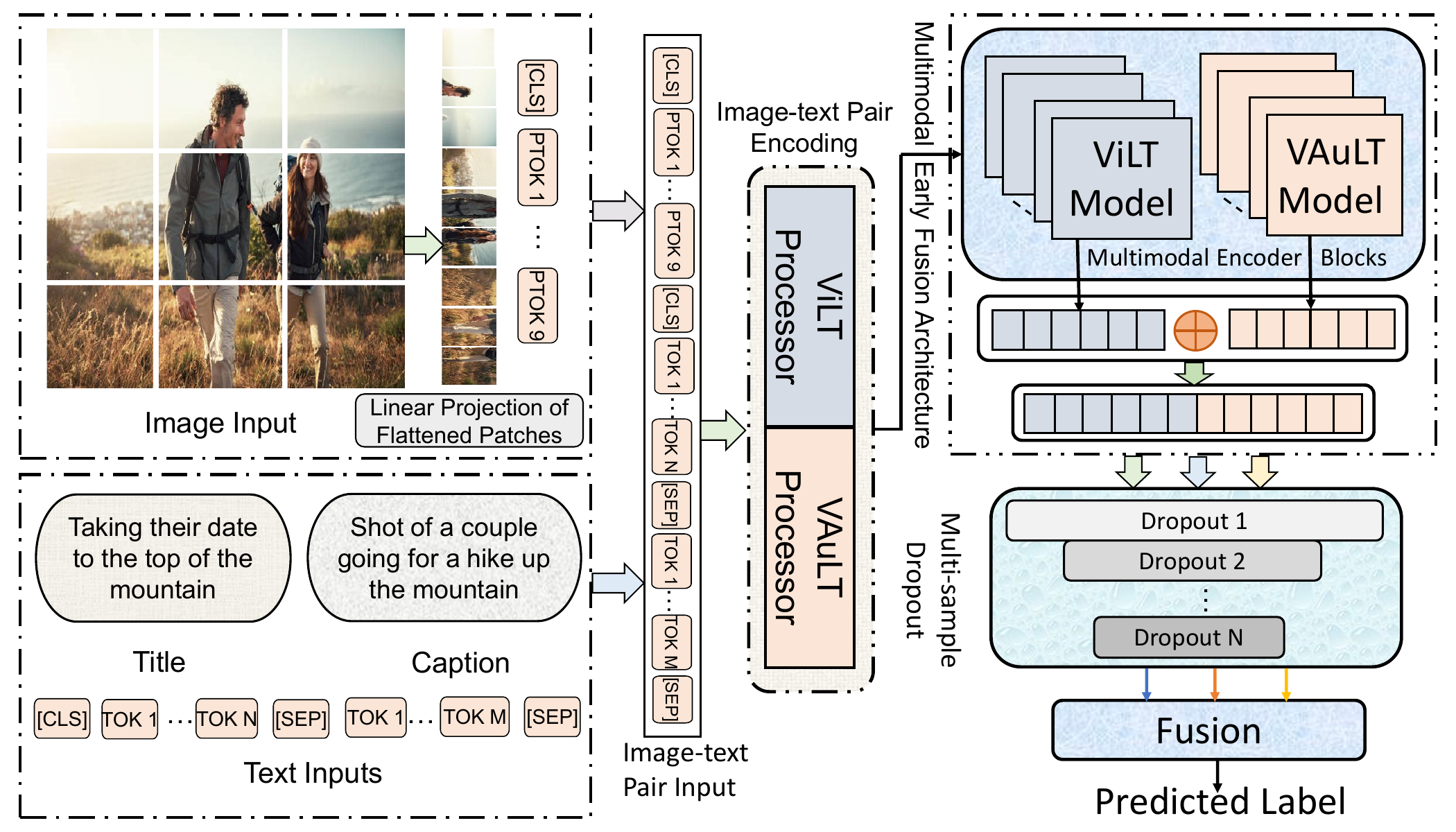}
\caption{Overview of our proposed framework for the human desire understanding task. To generate the integrated visual-contextualized feature vectors, the multimodal transformer model takes the input of the image-text pair. The concatenation of two model feature vectors creates the multimodal fusion architecture to get the diverse features of the task. Then, a classification head predicts the desire, sentiment, and emotion label.}
\label{fig:MOverview}
\end{figure*}
\section{Proposed framework}
\label{sec:proposedMethod}

Vision-and-language pre-training (VLP) transformers performed better in the vision-and-language multimodal task. To take advantage of VLP, we jointly fine-tuned two SOTA multimodal transformer models for the human desire understanding task. An overview of our proposed model for the human desire detection framework is depicted in~\figref{fig:MOverview}. The model is trained on the MSED dataset.

\textbf{Problem definition: } Given a text pair contains title $t$ and caption $c$ and a corresponding image $i$ of a tweet, classify the desire as $d$ of the tweet as $\epsilon$ $\{-family$, $romance$, $vengeance$, $curiosity$, $tranquility$, $social-contact$, $none\}$, emotion as $e$ $\epsilon$ $\{happiness$, $sad$, $neutral$, $disgust$, $anger$, $fear\}$, and sentiment as $s$ $\epsilon$ $\{positive$, $negative$, $neutral\}$. Given a training set \{ ($t_{1}$, $c_{1}$, $i_{1}$, $d_{1}$, $e_{1}$, $s_{1}$), . . . , ($t_{n}$, $c_{n}$, $i_{n}$$,$ $d_{n}$, $e_{n}$, $s_{n}$)\} of $n$ samples, our model’s objective is to maximize the function $f(\lambda_{d})$, $f(\lambda_{d})$, and $f(\lambda_{d})$ in Equation~\ref{eq:1} that take up a new instance denoted as $it_{k}$ (which contains $t_{k}$, $c_{k}$, $i_{k}$) to fitting desire $d_{k}$, emotion $e_{k}$, and sentiment $s_{k}$ labels, respectively. 

\begin{equation}
\label{eq:1}
\begin{aligned}
f(\lambda_{d})=\arg\max_{\lambda}\left(\prod_{k=0}^{n} P(d_{k} \mid (t_{k},c_{k}, i_{k})_{it};\lambda)\right)\\
f(\lambda_{e})=\arg\max_{\lambda}\left(\prod_{k=0}^{n} P(e_{k} \mid (t_{k},c_{k}, i_{k})_{it};\lambda)\right)\\
f(\lambda_{s})=\arg\max_{\lambda}\left(\prod_{k=0}^{n} P(s_{k} \mid (t_{k},c_{k}, i_{k})_{it};\lambda)\right)
\end{aligned}
\end{equation}

Where $it_{k}$ is the input texts-image pair whose desire label $d_{k}$, emotion label $e_{k}$, and sentiment label $s_{k}$ are to be predicted. $P$ denotes the log-likelihood function and the $\lambda$ symbol denotes the method's parameters we intend to optimize.

Given an input im$a$ge and texts are paired and transformed to the correct data formats, we feed them into the two multimodal transformer models' processors. We use two SOTA multimodal transformer models, ViLT and VAuLT, tuned on image-text pair settings to generate the image-text pair encoding. Each encoding is fed into the corresponding multimodal transformer model to get integrated visual-contextual feature vectors of each model. To get the diverse features representations of human desire understanding, we use a concatenation-based early fusion method on top of the two models' feature vectors. We jointly fine-tune this unified architecture to capture the domain-specific information of human desire understanding explicitly. To improve the generalizability and speed up the training of our proposed method, we use a multi-sample dropout training strategy. Then, a classification head is plugged on top of the multimodal unified architecture to predict the appropriate task-specific class label of the input image and texts.

\subsection{Multimodal transformer models}
\label{ref:transformerModels}
The BERT-based transformer model facilitates the learning of long-term dependency by handling the sentence as a whole rather than word by word. A combination of multi-head attention and positional embedding processes provides the necessary information regarding the relationship between words. To enable deep bidirectional representations of context, BERT uses masked language modeling. This model incorporates the feed-forward neural network architecture with a self-attention mechanism which helps distill contextual information more efficiently. In Vision Transformer (ViT), it divides an image into fixed-size patches, converts each patch into a linear embedding sequence, adds position embeddings, and feeds this sequence of vectors to a transformer encoder. It helps the ViT model to learn pixel-label information of the visual content effectively. Multimodal transformer models, including the ViLT and VAuLT models, use the unified architecture of the BERT-based text encoder and ViT image encoder in their backbone. Thus, it helps the model recognize the long-term dependency of context and establishes an intra-relationship between image and context. This relationship is critical for the human desire understanding task. We used two SOTA VLP transformer models, ViLT and VAuLT, to extract the integrated features vectors of the image-text pair for the human desire understanding task on social media data. Our multimodal approach for the human desire understanding task is based on ViLT as the backbone since it reflects the common design of multimodal transformers. Unlike other multimodal transformer models (VisualBERT), ViLT does not depend on modality-specific submodels for extracting features, and multiple objectives are used to pre-train the model, including Image Text Matching (ITM) and Masked Language Modeling (MLM).

\subsubsection{Vision-and-Language Transformer (ViLT)} 
The Vision-and-Language Transformer (ViLT)~\citep{kim2021vilt} processes visual inputs like text inputs, which implies a convolution-free method of processing visual data. The modal-specific components of ViLT are a class of VLPs that are less intensive to compute for multimodal interactions than the transformer component. ViLT is a parameter-efficient model that is much faster than other VLP models with region features. While it performs similarly or is even superior on downstream tasks, including VQA, image-text matching, multimodal classification, and multimodal entailment of vision-and-language, it is also at least four times faster than those models with grid features. To take advantage of the linear modality interaction, we use ViLT for the multimodal human desire understanding task. It helps ViLT to distill the relationships between image and text, which are crucial for the human desire detection task. To tokenize input text, ViLT uses the \emph{bert-base-uncased}~\footnote{\url{https://huggingface.co/bert-base-uncased}} tokenizer. To encode the input image, ViLT uses the ViT-B/32~\footnote{\url{https://huggingface.co/google/vit-base-patch32-224-in21k}} vison transformer model. It uses weights from ViT-B/32 pre-trained on ImageNet. It consists of 12 layers of transformer, with MLP size 3072, hidden size 768,  patch size 32, and 12 attention head layers. We use HuggingFace’s implementation of the \emph{vilt-b32-mlm}~\citep{kim2021vilt} checkpoint~\footnote{\url{https://huggingface.co/dandelin/vilt-b32-mlm}} of the ViLT model.

\subsubsection{Vision-and-Augmented-Language Transformer (VAuLT)}
The Vision-and-Augmented-Language Transformer (VAuLT)~\citep{chochlakis2022vault} expands the well-known VLP model ViLT. To improve the model's capacity to capture complex text features, the VAuLT model uses a modified version of the ViLT architecture that incorporates an extra text encoder module. Because ViLT was initialized from ViT, it lacks language understanding capabilities. VAuLT addresses this issue by substituting ViLT's language embedding inputs with linguistic features taken from a large LM pre-trained on more linguistic data varieties, which may also be chosen to better meet the demands of the downstream application. Training the LM and ViLT together in VAuLT improves the results over ViLT and achieves state-of-the-art or equivalent performance on various VL tasks incorporating richer language inputs and effective constructs. This motivates us to use the VAuLT model in our proposed framework for the human desire understanding task. That is why we use VAuLT along with ViLT to -capture the diversity of features. In VAuLT, we can use any language model to produce the sequence of embeddings which may be input into the ViLT architecture. To tokenize input text, we employ the \emph{bertweet-base} tokenizer. Due to the social media data used in MSED, the BERTweet may learn contextual information effectively. Thus, we use BERTweet to get diverse textual features representation in VAuLT~\footnote{\url{https://github.com/gchochla/VAuLT/tree/main/vault}} for human desire understanding task. We utilize HuggingFace’s implementation of the \emph{vilt-b32-mlm}~\citep{kim2021vilt} checkpoint of the ViLT model and \emph{vinai/bertweet-base}~\citep{nguyen2020bertweet} checkpoint~\footnote{\url{https://huggingface.co/vinai/bertweet-base}} as the language model in the VAuLT model training.

\subsection{Input representation of multimodal transformer model}
\begin{figure*}
\centering
\includegraphics[width=1\linewidth]{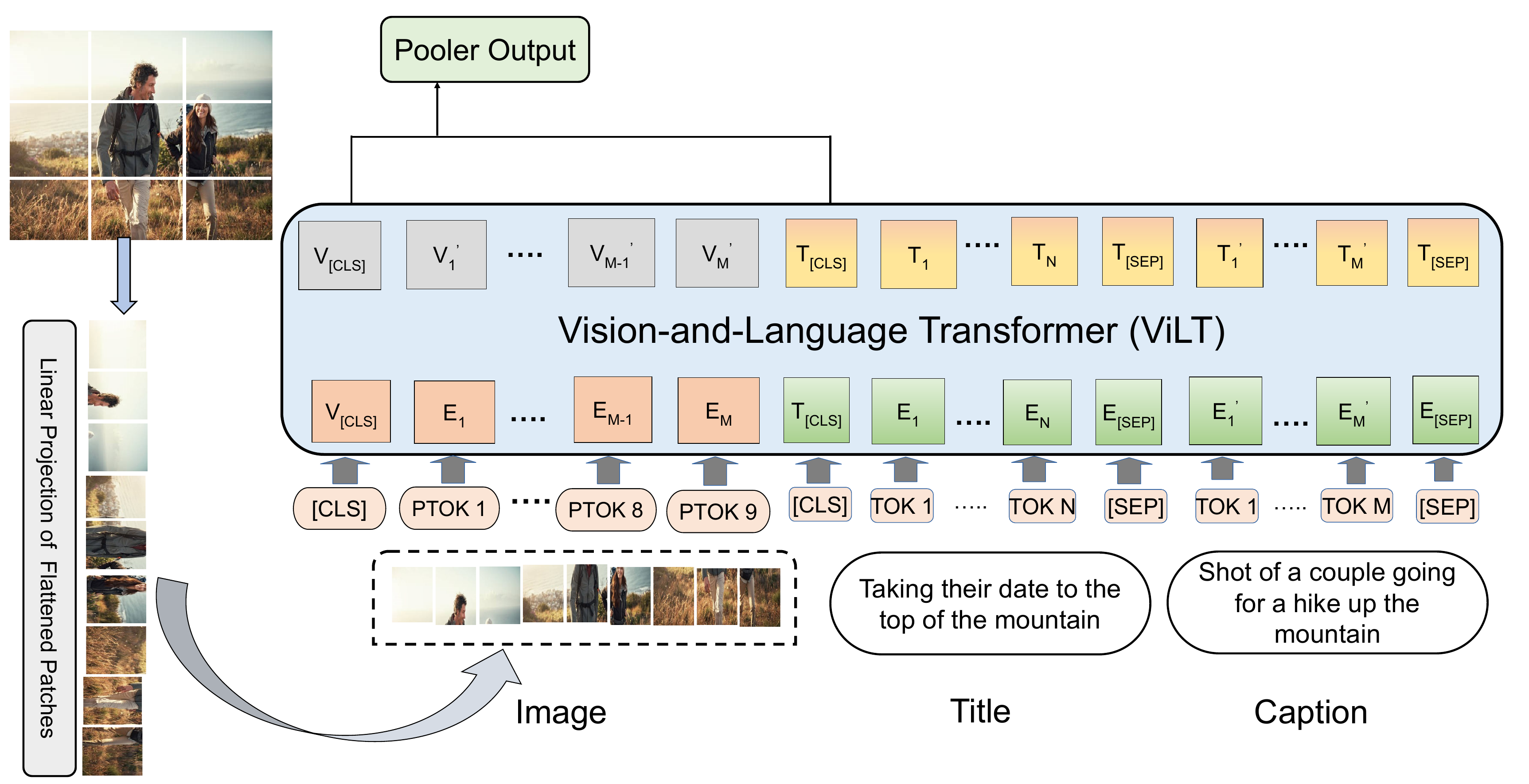}
\caption{Input representation of our proposed framework for the human desire understanding task.}
\label{fig:inputRepre}
\end{figure*}

In our proposed MMTF-DES method, we use two SOTA multimodal transformer models - ViLT and VAuLT. We now describe the input representation of these models. MSED contains two text inputs, a title and caption, and one visual input, a corresponding image. \figref{fig:inputRepre} illustrates the input representation used in our proposed method for representing the input data of MSED. We use an image-text pair training concept in the multimodal transformer models to train the human desire understanding task. The idea behind this concept is to combine the input image, title, and caption into one sequence in the multimodal transformer model's processor. Essentially, the caption and title pairs in the example are represented as a single sequence in which a special classification token [CLS] is inserted at the beginning of the first input (the caption), and a special separation token [SEP] is added between the caption and title to separate them. Therefore, the text-pair input sequence of multimodal transformer for our human desire understanding task is like ``[CLS] title [SEP] caption [SEP]" and the input image is sliced into patches where the patches denote as PTOK (patch token) and are flattened into the processor with a special classification token [CLS] at the beginning.

\subsection{Multimodal transformers early fusion}
The fusion of the models is an effective technique and common practice in multimodal learning that could improve the model's learning accuracy and robustness more than individual models. To benefit from the diversity of feature representations, we jointly fine-tune two SOTA VLP transformers, the ViLT and VAuLT models on the MSED dataset. In the case of multimodal models, an early fusion technique improves performance and generalizability. The advantage of early fusion is that it facilitates the exchange of information between the models at an earlier stage in the network, resulting in a better representation of the input data as a whole. We concatenate the pooler outputs of the ViLT and VAuLT models at an early stage, allowing the subsequent layers of the network to refine and combine these features in a more integrated and optimized manner. This enables us to get the two models integrated with visual-contextual feature vectors. With the fusion strategy, we have consolidated the diverse information of the human desire understanding task from these models. The same taxonomy of two multimodal transformer models with different text encoders improves the robustness and strength of our proposed model. This allows us to perceive the context and image pair from all aspects in a robust manner. 

\subsection{Multi-sample dropout training strategy}
With a multi-sample dropout-based training strategy, the base model is trained faster and generalizability is improved, leading to the enhanced overall performance of the system~\citep{inoue2019multi,hahn2020understanding,pan2021deepblueai}. During learning, the model is forced to learn a more generalizable representation that different sets of neurons are dropped out randomly for every input example. Therefore, it helps the base model avoid overfitting. To optimize the model, it calculates the loss for some dropout samples and averages them, resulting in a smaller loss. In our proposed method, the multi-sample dropout training strategy is used on top of the unified multimodal transformer-based architecture, where we employ a sample size of three dropout layers in the training process. In essence, we are duplicating the features vector of the unified architecture after the dropout layer is applied and sharing the weights among these duplicated layers which are fully connected. Training can be accelerated since more models are evaluated in the same time, resulting in faster convergence. To obtain the final loss, we aggregate the loss obtained from each sample and take their average. This way, the model can more quickly converge to a better model that generalizes well to new data as it learns more generalizable features. Later, a task-specific classification head on top predicts the appropriate desire, sentiment, and emotion labels for the corresponding image-text pair. 

\subsection{Fine-tuning procedure}
Fine-tuning multimodal transformer models may improve performance because it allows the model to be adapted to domain-specific knowledge. To achieve this, we jointly fine-tune the two multimodal transformer models described in Section~\ref{ref:transformerModels} using the domain-specific datasets. For diverse integrated contextual-visual feature vectors, a single concatenate layer is all we need. We train the human desire understanding task as an image-text-pair task where the title becomes the first text and the caption the second text in the input sequence. Additionally, the image is the third input of our proposed method. During fine-tuning, all other hyper-parameters remain the same as during the multimodal transformer model training, except for batch size, learning rates, epochs, max length, and dropout. The final hidden states of each model are then used to obtain a fixed-dimensional feature representation of the input sequence. Upon constructing the combined features vector of our method, we use an early fusion concatenation technique and then apply a multi-sample dropout technique to obtain the task-specific predicted class label.

\begin{equation}
\label{eq:2}
\begin{split}
T &= [t_{CLS};t_{1};....;t_{M};t_{SEP};c_{1};....;c_{M};t_{SEP}] \\ 
I &= [p_{CLS};p_{1};....;p_{M}]\\
V_{pool}&=(V_{m}(V_{pr}[T;I]))\\
VA_{pool}&=(VA_{m}(VA_{pr}[T;I]))\\
IFV&=V_{pool} \oplus VA_{pool}\\
O&=Avg.(OL(MSD(D_{0}(IFV))))
\end{split}
\end{equation}
Here, $T$ and $I$ denote the input text pair and image in Equation~\ref{eq:2}, respectively. $T$ contains two text inputs - the title and caption paired with each other with a special classification token, $t_{CLS}$ for text inserted at the beginning of the first input. $t_{1...M}$ represents the input title whereas $t_{i}$ represents each word in the title. The special separator token $t_{SEP}$ is in between the title and caption to separate them. Then, $c_{1...M}$ represents the input caption whereas $c_{i}$ denotes each word in the caption. $p_{1...M}$ represents the input image with a special classification token, $p_{CLS}$ for the image inserted at the beginning whereas $p_{i}$ indicates each patch in the input image. $T$ and $I$ are then passed into the corresponding model processor ($V_{pr}$ and $VA_{pr}$) to get the encoding of the text pair and image. Those encodings are then fed into the corresponding model ($V_{m}$ and $VA_{m}$) to extract the contextual-visual representation feature vector of the texts-image pair. In our proposed method, we use the pooler output of the ViLT and VAuLT models represented as $V_{pool}$ and $VA_{pool}$ in equations, respectively. To get the integrated features vector, $IFV$ we concatenate two feature vectors $V_{pool}$ and $VA_{pool}$. The $\oplus$  symbol denotes the concatenation in the equation. Later, $IFV$, is fed into a dropout layer, $D_{0}$ and the obtained output passes into the multi-sample dropout layer, $MSD$, to improve training. $MSD$ is then connected to the output layer, $OL$, and averages the obtained output to get the final prediction, $O$, of our method. This architecture is fine-tuned to learn task-specific knowledge by our proposed method.

\section{Experimental settings}
\label{sec:experimentalsettings}
\subsection{Dataset description}
To demonstrate the efficacy of our proposed multi-modal transformers fusion model, we evaluate our model on a publicly available multimodal benchmark dataset MSED~\citep{jia2022beyond}. MSED~\footnote{\url{https://github.com/MSEDdataset/MSED}} contains 9,190 text-image combinations gathered from various of social media, including Twitter, Getty Image, and Flickr. In the MSED dataset, the authors featured three multi-class classification tasks - sentiment analysis, emotion analysis, and desire analysis. In the sentiment analysis task, we need to determine positive, neutral, or negative sentiment classes in the text and corresponding image. Based on the image-text pair, we need to classify each instance as one of six classes - happiness, sad, neutral, disgust, anger and fear in the emotion analysis task. The six typical human desires - family, romance, vengeance, curiosity, tranquility, and social contact feature in the desire analysis task. The MSED dataset contains 6,127 training instances, 1,021 validation instances, and 2,024 test instances. The detailed statistics of the MSED dataset are presented in ~\tabref{tab:datasetCollection}.

\begin{table}[!htb]
\centering
\caption{Dataset statistics of MSED according to each label on the corresponding task.}
\begin{tabular}{p{3cm}|p{2.5cm}|ccc}
\toprule
Task   & Label & Train & Validation & Test \\
\midrule
\multirow{3}{*}{Sentiment Analysis} 
                           & Positive                  & 2,524  & 419        & 860  \\
                           & Neutral                   & 1,664  & 294        & 569  \\
                           & Negative                  & 1,939  & 308        & 613  \\
\midrule
\multirow{6}{*}{Emotion Analysis}   & Happiness                 & 2,524  & 419        & 860  \\
                           & Sad                       & 666   & 102        & 186  \\
                           & Neutral                   & 1,664  & 294        & 569  \\
                           & Disgust                   & 251   & 44         & 80   \\
                           & Anger                     & 523   & 78         & 172  \\
                           & Fear                      & 499   & 84         & 175  \\
\midrule
\multirow{7}{*}{Desire Analysis}    & Vengeance                 & 277   & 39         & 75   \\
                           & Curiosity                 & 634   & 118        & 213  \\
                           & Social-contract           & 437   & 59         & 138  \\
                           & Family                    & 873   & 152        & 288  \\
                           & Tranquility               & 245   & 39         & 87   \\
                           & Romance                   & 692   & 107        & 210  \\
                           & None                      & 2,969  & 507        & 1,031\\
\bottomrule
\end{tabular}
\label{tab:datasetCollection}
\end{table}

\subsection{Evaluation metrics}
To evaluate the system performance on the MSED dataset~\citep{jia2022beyond}, the authors used different strategies and metrics for the sentiment analysis, emotion analysis, and desire analysis tasks. For all three tasks of MSED, standard evaluation metrics, including Precision (P), Recall (R), and F1-score, are used to evaluate a system. To determine how well the model predicts a specific category in understanding human desire, the precision metric is used. The recall metric is used to understand how many times the model could detect a specific desire/emotion/sentiment category. The macro average category of those evaluation metrics was used because MSED is an imbalanced dataset that helps treat each class equally. The macro average F1-score is the primary evaluation measure in our analysis because it gives summarized information on both precision and recall metrics. We also report the results based on other evaluation measures too. 

\subsection{Model configuration}
\label{ref:modelConfig}
We describe the hyper-parameter settings for our proposed MMTF-DES model in this section. To enhance the efficiency of parallel computation of tensor image-text pairs of training, we used PyTorch to design and implement our proposed method. As part of this study, all experiments were carried out on the Google Colaboratory platform~\citep{bisong2019google}. The optimal hyper-parameters were selected using a grid search method based on the validation dataset. The hyper-parameters search space is illustrated in~\tabref{tab:hyperParameterSearch}.
\begin{table}[t]
\centering
\caption{The hyper-parameters search space of our proposed MMTF-DES method.\\}
\begin{tabular}{p{4cm}|l}
\toprule
Hyper-parameters  & Search Space\\
\midrule
Training batch size &$\lbrace 2, 4, 8, 16, 32\rbrace$\\[2pt]
\midrule
Test batch size &$\lbrace 1,2,4,8, 16\rbrace$\\[2pt]
\midrule
Max length &$\lbrace 32, 40, 64,128,256,512\rbrace$\\[2pt]
\midrule
Learning rate &$\lbrace 1e-3, 1e-5, \cdots, 5e-6\rbrace$\\[2pt]
\midrule
Number of epochs   & $\lbrace 1, 2, 3, \cdots, 9, 10\rbrace$\\[2pt]
\midrule

Dropout  &$\lbrace 0.1, 0.2, \cdots, 0.8\rbrace$\\
\bottomrule
\end{tabular}
\label{tab:hyperParameterSearch}             
\end{table}

We used two pre-trained multimodal transformer models~\citep{wolf2020transformers}, VAuLT and ViLT, as multimodal encoders to extract effective integrated contextual-visual features in our proposed MMTF-DES model. Prior research~\citep{aziz2021csecu} reveals that fine-tuning the hyper-parameters of those models makes them consistently outperform the pre-trained models for downstream tasks. Several hyper-parameters, including training batch size, test batch size, learning rate, dropout, and epochs, were fine-tuned to get the optimal value. We train our proposed method with the provided training data with the validation data for efficient training. We trained VAuLT and ViLT models using 5 epochs and set the learning rate to nearly the same as 3e-5. The best settings of these hyper-parameters are reported in~\tabref{tab:hyperParameter}. We used the \emph{vilt-b32-mlm} checkpoint of the ViLT model and \emph{vilt-b32-mlm} and the \emph{vinai/bertweet-base} checkpoint for the VAuLT model. We used the CUDA-enabled GPU and set the fixed manual seed to generate reproducible results. During training, we saved our model based on the best validation loss by evaluating the validation set. To reduce the noisy features and avoid overfitting, we fine-tuned the dropout rate hyper-parameter. 

\begin{table}[t]
\centering
\caption{The optimal value of hyper-parameters used in our proposed MMTF-DES system.\\}
\begin{tabular}{p{4cm}|l|l|l}
\toprule
Hyper-parameters  & Sentiment Analysis & Emotion Analysis & Desire Analysis\\
\midrule
Training batch size &4  &8 &8\\
Test batch size &1 &1 &1\\
Learning rate &3e-3 &2.99e-3 &3.1e-3 \\
Max length &40 &40 &40\\
Dropout  &0.5 &0.5 &0.7 \\
Multi-sample dropout &0.1, 0.2, 0.3 &0.1, 0.2, 0.3 &0.1, 0.2, 0.3 \\
Number of epochs   &5 &5 &5 \\
\bottomrule
\end{tabular}
\label{tab:hyperParameter}             
\end{table}

\section{Experimental results and analysis}
\label{sec:experimentsAndEvaluation}
We now evaluate the performance of our proposed MMTF-DES approach. The objectives of our experimental design are six-fold: (1) we analyze different modality baseline models to choose the best model (2) we examine the performance of our used fusion strategies to pick the most effective one that we used in all the later experiments (\textbf{RQ2}); (3) we analyze the overall performance and task-wise performances of our proposed method on the MSED dataset; and (4) we determine the performance of individual multimodal transformer models used in our proposed approach (\textbf{RQ1}). (5) We proposed a new desire analysis task where we describe the task and evaluate our method on this task (\textbf{RQ5}). (6) We provide a comparative performance analysis between our proposed method and other current SOTA (\textbf{RQ3}).

\subsection{Analysis of different modality baseline models}
We present different modality baseline methods analysis on the human desire understanding task that drove this work. We consider three categories of modality in the baseline analysis discussion. From the textual modality, we consider the BERTweet transformer model since text data are annotated from tweet data. In the visual modality, we consider the ResNet model as it has outstanding performance in many image classification tasks. In the multimodal category, we use the ViLT multimodal transformer model for the human desire understanding task to take advantage of its linear modality interaction (text and image modality treated with the same priority).  

\begin{table}[!htb]
\centering
\caption{Performance of different modality baseline models on the MSED dataset.\\}
\begin{tabular}{p{3cm}|p{2cm}|p{2cm}|ccc}
\toprule
Task   & Method & Modality & Precision & Recall& Macro F1-score \\
\midrule
\multirow{4}{*}{Sentiment Analysis} 
                           &BERTweet &Text &82.25 &83.62 &82.49\\
                           & ResNet  &Image & 70.85  &70.61  & 70.64  \\ 
                           &ViLT &Multimodal &85.62 &86.14 &85.81\\
\midrule
\multirow{4}{*}{Emotion Analysis} 
                           &BERTweet &Text &80.99 &77.15 &78.34\\
                           & ResNet &Image  & 58.74  & 54.67  & 56.40  \\
                           &ViLT &Multimodal &79.17 &83.09 &80.81\\
\midrule
\multirow{4}{*}{Desire Analysis} 
                           & BERTweet &Text &77.11 &81.19 &78.86\\                           
                           & ResNet   &Image& 49.97  & 49.35 & 49.20  \\  
                           & ViLT &Multimodal  &81.23 &75.16 &77.78 \\
\bottomrule
\end{tabular}
\label{tab:baselineModaleResult}
\end{table}

The results of the various modality baseline experiments are presented in Table~\ref{tab:baselineModaleResult}. To show the baseline methods performance on the MSED dataset, we show the performance of all three subtasks - sentiment analysis, emotion analysis, and desire analysis. In sentiment analysis, the BERTweet model achieves  82.49, the ResNet model 70.64, and the ViLT model 85.81 in terms of the primary evaluation measure macro-averaged F1 score. This indicates for the multimodal transformer model, the ViLT performance is 3.87\% and 17.68\% higher than the BERTweet and ResNet model performance, respectively. In emotion analysis, the BERTweet model achieves 78.34, the ResNet model 56.40, and the ViLT model 80.81 in terms of the primary evaluation measure macro-averaged F1 score. This indicates for the multimodal transformer model, the ViLT performance is 3.05\% and 30.2\% higher than the BERTweet and ResNet model performance, respectively. In desire analysis, the BERTweet model achieves 78.86, the ResNet model 49.2, and the ViLT model 77.78 in terms of the primary evaluation measure macro-averaged F1 score. This indicates for the multimodal transformer model, the ViLT gives a similar performance to the BERTweet and a 36.75\% higher score than the ResNet model performance. This validates the significance of the multimodal transformers model in the human desire understanding task.

\begin{table}[!hb]
\centering
\caption{Performance (Macro F1-scores; higher is better) of our used fusion strategies on the MSED dataset. \\ }
\begin{tabular}{@{}l@{\hspace*{0.5cm}}| c@{\hspace*{0.5cm}}| c@{\hspace*{0.5cm}}}
\toprule
Task  & \makecell{Early Fusion\\(Macro F1-score)} & \makecell{Late Fusion\\(Macro F1-score)}\\
\midrule
Sentiment Analysis & 88.44 &87.39\\
\midrule
Emotion Analysis & 84.26 &79.96\\
\midrule
Desire Analysis &83.11 &78.47\\
\bottomrule
\end{tabular}
\label{tab:compareIntegrationStrategy}             
\end{table}

\subsection{Performance of fusion techniques}
\label{sec:rq2}
To identify the best fusion technique for our proposed MMTF-DES, we apply two categories of fusion techniques - early fusion and late fusion~\citep{snoek2005early}. In early fusion, we concatenate the ViLT and VAuLT multimodal transformer models' feature vectors together at an early stage of the network, typically after the initial feature extraction layers of each model. In concatenate-based late fusion, the feature vectors from the two multimodal transformer models are combined later in the network, typically after passing through various dense layers. That means each model feature vector is fed into many dense layers and concatenated together. In these two fusion strategies, the early fusion concatenation strategy achieved the best performance across all three tasks. \tabref{tab:compareIntegrationStrategy} illustrates the impact of both fusion strategies based on the primary evaluation metrics macro average F1-score across all three tasks (\textbf{RQ2}). Here, early fusion results in a better feature representation of the input image-text pair data, which helps the model improve the performance of the human desire understanding task. It enables the model to capture more complex and subtle relationships in the image-text pair data, which can improve the model's ability to generalize to new and unseen data. Thus, the network learns a more integrated and optimized set of features by fusing the features at an earlier stage in the network. In the sentiment analysis task, the early fusion-based approach achieved an 88.44\% macro average F1-score, which is 1.2\% higher than the late fusion-based model's-performance. The early fusion-based method shows significant performance improvement over the late fusion-based method on the emotion analysis and desire analysis tasks, with a 5.1\% and 5.6\% higher macro average F1-score, respectively. It validates the efficacy of the early fusion-based strategy. Therefore, we choose this
fusion strategy for our proposed MMTF-DES method and the rest of the results are reported following this setting.

\begin{table}[ht]
\centering
\caption{Performance of our proposed MMTF-DES model for the MSED dataset.}
\label{tab:modelResult}   
\resizebox{.7\textwidth}{!}{\begin{tabular}{p{4cm}cccc}
\toprule
Task & Precision &  Recall  & Macro F1-score & Accuracy\\
\midrule
Sentiment analysis  &88.27     &88.68     &88.44 &88.44\\
Emotion analysis     &84.39    &84.64     &84.26 &84.13\\
Desire analysis     &84.23     &82.01     &83.11 &86.97\\
\midrule
Average             &85.63     &85.11     &85.27 &86.52\\
\bottomrule
\end{tabular}
}
\end{table}

\subsection{Overall performance across three tasks}
The summarized results of our proposed MMTF-DES method based on different tasks are presented in~\tabref{tab:modelResult}. The overall performance for the MSED dataset is 85.27\% and 86.52\% based on the macro averaged F1-score and accuracy score, respectively. Here, our proposed method performs better for the sentiment analysis task than for the emotion analysis and desire analysis tasks. A possible reason for this difference in performance may be related to the nature of the three tasks. Sentiment analysis is concerned with identifying the overall sentiment or opinion expressed in text and images. In contrast, emotion analysis and desire analysis are concerned with identifying the specific emotional states and desires expressed by an individual that are inherently embedded in images and text. Therefore, sentiment analysis may be a more straightforward task that is easier to model, while emotion analysis and desire analysis may be more complicated and nuanced.

\begin{table}[t]
\centering
\caption{Individual component performance of our proposed MMTF-DES model for the MSED dataset.}
\label{tab:IndividualmodelResult}   
\resizebox{.7\textwidth}{!}{\begin{tabular}{p{4cm}cccc}
\toprule
\multicolumn{1}{c}{Method} & \multicolumn{1}{c}{Precision} &  \multicolumn{1}{c}{Recall}  &\multicolumn{1}{c}{Macro F1-score}& \multicolumn{1}{c}{Accuracy}\\
\midrule
\multicolumn{5}{l}{\textit{Sentiment analysis task}}\\
\midrule
VAuLT &84.61 &85.65 &84.94 &85.16\\
ViLT &85.62 &86.14 &85.81 &86.04\\
MMTF-DES (proposed)  & \textbf{88.27} &\textbf{88.68} &\textbf{88.44} &\textbf{88.44}\\
\midrule
\multicolumn{5}{l}{\textit{Emotion analysis task}}\\
\midrule
VAuLT &\textbf{84.56} &79.39 &81.47 &83.21\\
ViLT &79.17 &83.09 &80.81 &81.29\\
MMTF-DES (proposed)  &84.39 &\textbf{84.64}    &\textbf{84.26} &\textbf{84.13}\\
\midrule
\multicolumn{5}{l}{\textit{Desire analysis task}}\\
\midrule
VAuLT &80.33  &\textbf{83.32} &81.45 &85.55\\
ViLT &81.23 &75.16 &77.78 &83.05\\
MMTF-DES (proposed) &\textbf{84.23}     &82.01     &\textbf{83.11} &\textbf{86.97}\\
\bottomrule
\end{tabular}
}
\end{table}

\subsection{Impact of individual multimodal transformer models}
\label{sec:rq1}
We further analyze the performance of our proposed MMTF-DES method by evaluating the performance of individual multimodal transformer models. We only retain one multimodal transformer model at a time and discard the other model to do this. To examine the component analysis of our proposed method, we used the MSED dataset for all three tasks, and the evaluation results are illustrated in~\tabref{tab:IndividualmodelResult}. 
To capture diverse integrated contextual-visual representation, we incorporate two multimodal transformer models - ViLT and VAuLT. We apply an early fusion at the very early stage of the network i.e., the feature's label fusion. Such fusion is crucial for learning pixels-based information in visual content and semantic information in context from two modalities in an integrated manner. Jointly trained our proposed unified multimodal architecture  performed better than the individual models across all three tasks (\textbf{RQ1}). In the sentiment analysis task, the ViLT model performs better than the VAuLT model. However, MMTF-DES shows a 2.97\% performance improvement over ViLT. The VAuLT model performs better than the ViLT model in the emotion analysis and desire analysis tasks. Nevertheless, our proposed MMTF-DES method achieved a 3.32\% and 2.0\% higher macro averaged F1-score on the emotion analysis and desire analysis tasks, respectively. This highlights the importance of using diverse multimodal transformer encoders in our proposed method.
\begin{table}[!h]
\centering
\caption{The statistics of our proposed binary-type desire analysis task data of the MSED dataset.}
\begin{tabular}{@{}l@{\hspace*{0.7cm}}| c@{\hspace*{0.7cm}}| c@{}}
\toprule
Data Category  & Desire & Not Desire\\
\midrule
Train &3158  &2969\\
\midrule
Dev &514  & 507\\
\midrule
Test &1011 &1031\\
\bottomrule
\end{tabular}
\label{tab:binaryDesiredata}             
\end{table}

\subsection{Binary desire analysis}
\label{sec:rq5}
To further extend this task, we propose a new task for the human desire understanding task (\textbf{RQ5}). It is important to work on the human desire analysis task - we first need to identify whether the image-text pair contains human desire. Then, if the pair contains desire, the next task is to identify which type of desire; prior research follows this trend~\citep{jin2022automatic}. However, the MSED dataset is not designed as a binary desire analysis task together with the existing multiclass desire analysis task. Thus, we propose a new desire analysis task as a binary classification. To do this, we consider the \emph{Not Desire} and \emph{Desire} class labels in this task for the image-text pair. In the \emph{Desire} class label, we use vengeance, curiosity, social contract, family, tranquility, and romance label data as \emph{Desire} class data because authors~\citep{jia2022beyond} consider \emph{Desire} image-text pair data into these six categories, for \emph{Not Desire} data we consider the same as authors provided (\emph{None} class label data of desire analysis task in~\tabref{tab:datasetCollection}). According to this, we distribute those data in the binary label. Our proposed binary-type desire data statistics are presented in~\tabref{tab:binaryDesiredata}.

To know how well our proposed method identifies desire image-text pair data, we train our proposed method for the binary desire classification task. For binary desire classification, our proposed MMTF-DSE method achieved 90.21\% precision, 90.19\% recall, and a 90.21\% macro averaged F1-score. Although, our proposed method achieved an 83.11\% macro average F1-score in the multiclass desire analysis task. A higher macro F1-score demonstrated the applicability and generalizability of our proposed early fusion of a multimodal transformers-based approach for the human desire understanding tasks.

\subsection{Comparative performance analysis}
\label{sec:rq3}
\subsubsection{Comparing with state-of-the-art study on MSED}
To evaluate the performance of our proposed method against the current state-of-the-art techniques, we compared the performance with the top-performing systems on the MSED dataset. The authors proposed three tasks - desire analysis, sentiment analysis, and emotion analysis - to evaluate the MSED dataset and present various strong baselines by combining diverse features fusion~\citep{jia2022beyond}. They used two categories of encoders to extract the text and image features, respectively. To represent the text, they use three well-known text encoders - deep CNN (DCNN), bidirectional LSTM (BiLSTM)~\citep{zhang2020quantum}, and the pre-trained language model BERT~\citep{devlin2019bert}. To encode images, they use two widely used visual encoders, i.e., AlexNet~\citep{alom2018history} and ResNet~\citep{szegedy2017inception}. Among their modality analysis, the BERT model achieved the best macro averaged F1-score in the textual modality and  ResNet achieved better performance in the visual modality. The fusion of BERT and ResNet learns the visual and textual information effectively and achieves the best performance among all modalities. Also, they used a state-of-the-art multimodal pre-trained model, Multimodal Transformer~\citep{gabeur2020multi}, as one of the baselines. It performs better than other models except for the fusion of the BERT-ResNet model. Recently, M3GAT~\citep{zhang2023m3gat} proposed a graph attention network-based method focusing on the sentiment and emotion analysis task. However, they did not report their result on the desire analysis task.
\begin{table}[t]
\centering
\caption{Comparative performance (precision (P), recall (R), Macro F1-score, and accuracy: higher is better) analysis of our proposed MMTF-DES model with the other SOTA method on MSED dataset. $\nabla$ denotes an improvement of our proposed method against current SOTA.}
\label{tab:ComparativeResult}   
\resizebox{.8\textwidth}{!}{\begin{tabular}{p{7cm}ccc}
\toprule
\multicolumn{1}{c}{Method} & \multicolumn{1}{c}{Precision} &  \multicolumn{1}{c}{Recall}  &\multicolumn{1}{c}{Macro F1-score}\\
\midrule
\multicolumn{4}{l}{\textit{Sentiment analysis task}}\\
\midrule

Multimodal Transformers~\citep{jia2022beyond} &83.56 &83.45 &83.50 \\
M3GAT~\citep{zhang2023m3gat}  &84.66 &85.15 &84.85\\
BERT+ResNet~\citep{jia2022beyond} &85.83 &85.79 &85.81\\
MMTF-DES (proposed)  & \textbf{88.27} &\textbf{88.68} &\textbf{88.44}\\
$\nabla$ SOTA & & &(+3\%)\\
\midrule
\multicolumn{4}{l}{\textit{Emotion analysis task}}\\
\midrule
Multimodal Transformers~\citep{jia2022beyond} &81.62 &81.61 &81.53\\
M3GAT~\citep{zhang2023m3gat}  &82.53 &81.51 &81.97\\
BERT+ResNet~\citep{jia2022beyond} &83.54 &81.51 &82.42\\
MMTF-DES (proposed) &\textbf{84.39} &\textbf{84.64}    &\textbf{84.26}\\
$\nabla$ SOTA & & &(+2.2\%)\\
\midrule
\multicolumn{4}{l}{\textit{Desire analysis task}}\\
\midrule
Multimodal Transformers~\citep{jia2022beyond} &81.42 &80.20 &80.92 \\
BERT+ResNet~\citep{jia2022beyond} &83.43 &\textbf{82.43} &82.28\\
MMTF-DES  (proposed) &\textbf{84.23}     &82.01    &\textbf{83.11}\\
$\nabla$ SOTA & & &(+1\%)\\
\bottomrule
\end{tabular}
}
\end{table}

\subsubsection{Comparative performance across all three tasks}
To evaluate the performance of our proposed method against the current SOTA, we compared the performance with the top-performing systems (\textbf{RQ3}) on the MSED dataset. We fuse two pre-trained multimodal transformer models - ViLT and VAuLT - to get the unified multimodal architecture for the human desire understanding task. To compare the performance of our proposed MMTF-DES model, we used the MSED dataset, which contains three tasks - sentiment analysis, emotion analysis, and desire analysis. The comparative performance of our proposed MMTF-DES system on test data against other SOTA systems is presented in~\tabref{tab:ComparativeResult}. Our proposed multimodal early fusion-based transformer approach outperforms other SOTA models across all three tasks (\textbf{RQ3}). Our proposed method achieves 88.44\% on the sentiment analysis task, 84.26\% on the emotion analysis task, and 83.11\% on the desire analysis task, based on the primary evaluation metric macro average F1-score on the MSED dataset. Our proposed MMTF-DES model outperforms the prior best-performing fusion of the BERT-ResNet model by 3\% on the sentiment analysis task, 2.2\% on the emotion analysis task, and 1\% on the desire analysis task (as denoted by $\nabla$ in the ~\tabref{tab:ComparativeResult}). The comparative performance analysis confirms that an approach that involving the fusion of various multimodal transformers encoders can achieve good performance for the human desire understanding task from image-text across different associated tasks. This validates the effectiveness and applicability of our proposed method on the multimodal human desire understanding task.

Those models do not perform as well as our proposed method because they struggle to achieve intra-and-inter relationships between modalities as they fuse different modalities externally. To learn the intra-and-inter relationship between modalities, the pairwise training of different data, including visual and textual, is crucial for the desire understanding task. To mitigate this issue, we use multimodal transformer models with pairwise training of the text and image. This helps us achieve the intra-relationship between image and text. We fuse two multimodal transformer models, to extract diverse integrated visual-contextual features representation which helps us to achieve the inter-relationship of two modalities. The early fusion of two multimodal transformer models results in a better feature representation of the input image-text pair data, which helps the model to improve the performance of the human desire understanding tasks. It enables the model to capture more complex and subtle relationships in the image-text pair data, which can help improve the model's generalizability to unseen data. Moreover, we use a multi-sample dropout training strategy to improve the generalizability of our proposed model.

\section{Discussion}
\label{sec:discussion} 
In this section, we proffer some research findings into the multimodal human desire understanding research. We provide modality dominance analysis, multi-sample dropout impact, research analysis, and error analysis to analyze the efficacy of our proposed MMTF-DES method in the human desire understanding domain.

\subsection{Modality dominance}
To analyze the modality dominance in the multimodal human desire understanding task, we discuss the individual modality model's result across all three tasks. \tabref{tab:modalityResult} shows the results of individual model performance on the MSED dataset. Here, we consider three text encoders - BiLSTM, BERT, and BERTweet, and two Visual encoders including AlexNet, and ResNet in our modality dominance comparison. Across modalities, we have seen that all text models perform better than all image models by a large margin. Hence, the multimodal human desire understanding task is a text modality-dominant task. This research finding is critical for working on the multimodal human desire understanding task. This modality dominance experiment motivates us to use the same taxonomic two multimodal models with different text encoders. In our proposed method, we use two multimodal transformer models - ViLT and VAuLT - in a unified architecture for the human desire understanding task. The VAuLT model used the same architecture as the ViLT model but generalized the text encoder. ViLT employs BERT as a text encoder in the model backbone. Because of text modality dominance in the human desire understanding task, we try to get diversity in text modality. Hence, we used BERTweet as a text encoder in VAuLT, as the MSED dataset used social media data (Twitter data too).The BERTweet text encoder is pre-trained on Twitter data, which may help the model learn semantic information in a robust way.
\begin{table}[!htb]
\centering
\caption{Individual modality models performance on the MSED dataset.\\}
\begin{tabular}{p{3cm}|p{2.5cm}|ccc}
\toprule

Task   & Method & Precision & Recall& Macro F1-score \\
\midrule
\multirow{5}{*}{Sentiment Analysis} 
                          & BiLSTM (Text)  & 78.43  &78.75 & 78.58  \\
                           & BERT (Text)  & 84.43  &84.28  & 84.35  \\
                           &BERTweet (Text) &82.25 &83.62 &82.49\\
                           & AlexNet (Image)  & 68.76  &68.21  & 68.45  \\
                           & ResNet  (Image) & 70.85  &70.61  & 70.64  \\                           
\midrule
\multirow{5}{*}{Emotion Analysis} 
                          & BiLSTM (Text) & 73.49  & 72.17 & 72.73  \\
                           & BERT  (Text) & 81.76  & 80.57 & 81.10  \\
                           &BERTweet (Text) &80.99 &77.15 &78.34\\
                           & AlexNet (Image)  & 56.42  & 53.29 & 54.66  \\
                           & ResNet (Image)  & 58.74  & 54.67  & 56.40  \\                           
\midrule
\multirow{5}{*}{Desire Analysis} 
                          & BiLSTM   (Text) & 73.20  & 67.82 & 69.14  \\
                           & BERT   (Text) & 81.74  & 80.39 & 80.88 \\
                           & BERTweet (Text) &77.11 &81.19 &78.86\\                           
                           & AlexNet   (Image)& 51.47  & 49.33  & 50.07  \\
                           & ResNet   (Image)& 49.97  & 49.35 & 49.20  \\                           
\bottomrule
\end{tabular}
\label{tab:modalityResult}
\end{table}

\subsection{Impact of multi-sample dropout}
\label{sec:rq4}
To analyze the impact of multi-sample dropout (MSD) in our proposed method, we experiment with the MSED dataset. In this experiment, we remove the multi-sample dropout module from our proposed MMTF-DES method to discover the significance of the MSD in the human desire understanding task. The results without MSD and with MSD (our proposed method) are reported in~\tabref{tab:MSDImpact}. We have seen that the method without MSD results in significantly lower performance than our proposed method (with MSD) across all three tasks in terms of the macro averaged F1-score. With the MSD, our proposed method improves the performance by 1.32\% on the sentiment analysis task, 1.72\% on the emotion analysis task, and 2.22\% on the desire analysis task, based on the primary evaluation measure macro averaged F1-score (\textbf{RQ4}). This highlighted the importance of adding multi-sample dropout layers to the unified multimodal transformer-based model architecture.

\begin{table}[hbt]
\centering
\caption{Performance on the MSED dataset with (our proposed MMTF-DES method) and without (w/o) multi-sample dropout (MSD).\\}
\begin{tabular}{p{3cm}|p{2.5cm}|ccc}
\toprule

Task   & Method & Precision & Recall& Macro F1-score \\
\midrule
\multirow{3}{*}{Sentiment Analysis} 
                          & MMTF-DES  & \textbf{88.27} &\textbf{88.68} &\textbf{88.44}\\
                           & W/o MSD  & 87.12  &87.45  & 87.27  \\
                   
\midrule
\multirow{3}{*}{Emotion Analysis} 
                          & MMTF-DES &\textbf{84.39} &\textbf{84.64}    &\textbf{84.26}\\
                           & W/o MSD   & 83.15  & 83.14 & 82.81  \\
                           
\midrule
\multirow{3}{*}{Desire Analysis} 
                          & MMTF-DES   &\textbf{84.23}     &82.01    &\textbf{83.11}\\
                           & W/o MSD  & 81.19  & \textbf{82.34} & 81.27 \\
                        
\bottomrule
\end{tabular}
\label{tab:MSDImpact}
\end{table}

\subsection{Research analysis}
\label{sec:researchfindings}
We presented five research questions in Section~\ref{sec:rqs} that motivated us to perform the human desire understanding research. The first question (\textbf{RQ1}) concerned capturing better integrated visual-contextual features from different image-text pair data. We used two multimodal transformer encoders - ViLT and VAuLT - to extract diverse visual-contextual features. The analysis results of Section~\ref{sec:rq1} demonstrated the efficacy of exploiting diverse multimodal transformer encoders in the human desire understanding task. In the second question (\textbf{RQ2}), we focused on effective fusion techniques for leveraging the above-mentioned multimodal transformer models to get a unified multimodal architecture. We showed the comparative performance between the early fusion-based concatenation and late fusion-based concatenation techniques in Section~\ref{sec:rq2}. The early fusion-based concatenation achieved a macro averaged F1-score 3.97\% higher than the late fusion-based concatenation technique on the MSED dataset for our proposed method. Next, in the third question (\textbf{RQ3}), we investigated a comparative performance analysis of our proposed MMTF-DES method with other state-of-the-art methods (i.e., BERT-ResNet, Multimodal transformer, etc.), as described in Section~\ref{sec:rq3}. The comparative analysis based on the MSED dataset demonstrates that our proposed MMTF-DES method achieved a 3\% improvement on the sentiment analysis task, a 2.22\% improvement on the emotion analysis task, and a $\sim$1\% improvement on the desire analysis task compared to the existing best-performing model fusion of BERT and ResNet method. This model struggles to learn the inter and intra-relational structure of the image-text pair relationship to identify human desire. To improve the base model performance, we use a training strategy called multi-sample dropout. It improves the base method performance by an average of 1.7\% on the MSED dataset (\textbf{RQ4}). The performance analysis in Section~\ref{sec:rq4} demonstrated the efficacy of using multi-sample dropout in our proposed MMTF-DES method. Moreover, we propose a new binary desire analysis task to effectively understand human desire analysis (\textbf{RQ5}). The task description and performance of our method on this task are presented in Section~\ref{sec:rq5}   

\begin{figure*}[!ht]
\centering
\includegraphics[width=1\linewidth]{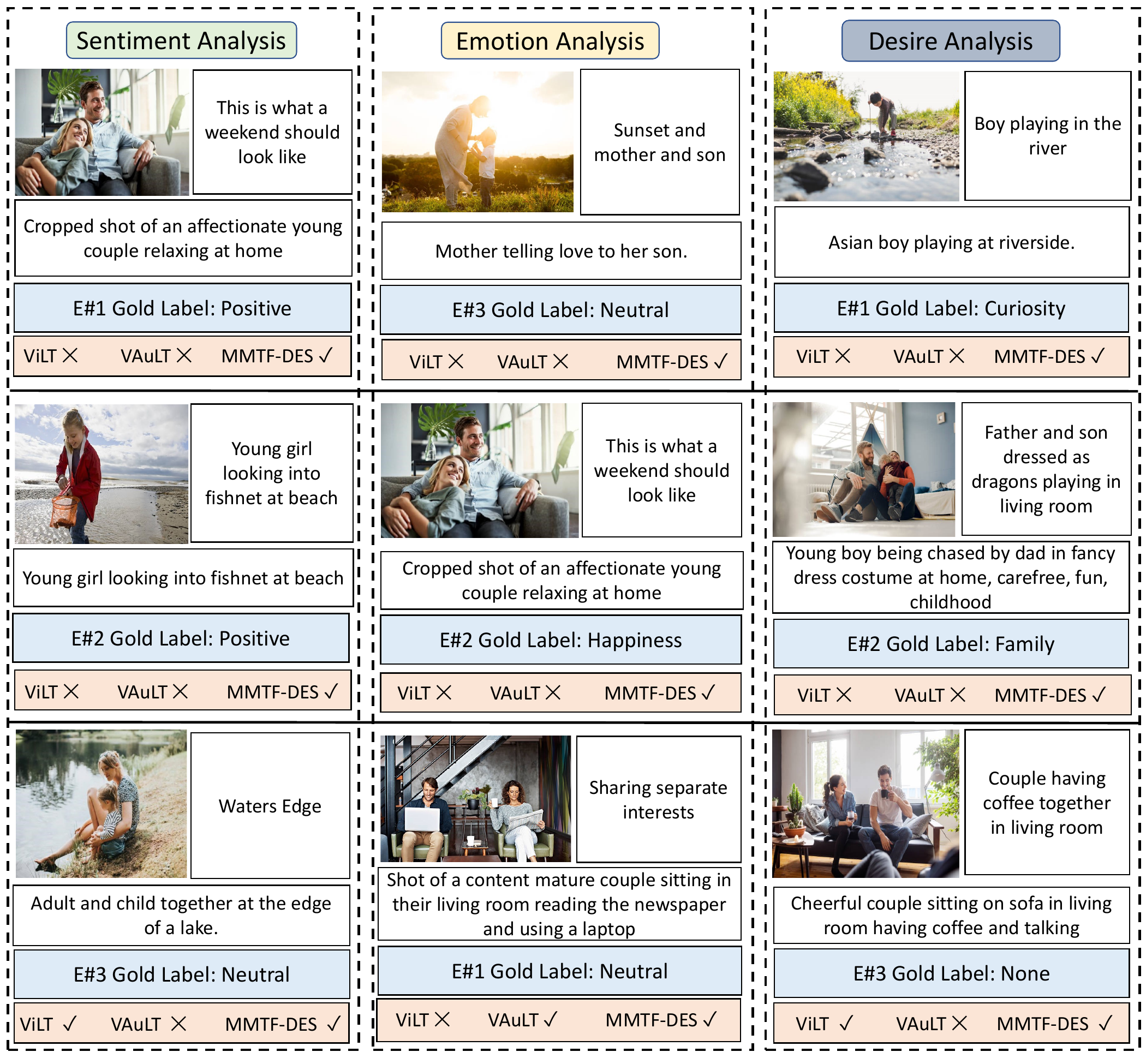}
\caption{Performance analysis and error analysis of our proposed framework for the human desire understanding task. A \protect\checkmark indicates correctly classified or matched with the gold label and a \protect\crossmark~ represents erroneously classified.}
\label{fig:ErrorRepre}
\end{figure*}

\subsection{Performance analysis and error analysis}
To enhance understanding of the potential of our proposed method, we conduct error and performance analyses. To do this, we articulate a few multimodal error cases that were misclassified by the individual model and successfully classified by our proposed method across all three tasks from the MSED dataset, as depicted in~\figref{fig:ErrorRepre}. In this figure, we show the predictions of three models, ViLT, VAuLT, and MMTF-DES where our proposed MMTF-DES model's predicted labels matched with the gold label for each task and individual component (ViLT and VAuLT) were predicted erroneously. Across all three tasks, our system correctly predicts the class label in E\#1 and E\#2, even if both individual components fail to predict the correct label. This is because with diverse features of multiple models, early fusion helps the model to learn the image-text pair relationship effectively. Hence, our proposed method correctly predicts the class label where individual components predict erroneously. However, in example E\#3 of all tasks, our approach correctly classifies the class label; the ViLT model also predicts the correct label for sentiment analysis and desire analysis, for example, E\#3; for emotion analysis, the VAuLT model predicts the correct class label. Here, the individual models help the model to obtain the correct predicted label; if one individual model fails to predict the correct label, the other model helps the overall model to predict the correct class label. This highlights the effectiveness of using multiple multimodal transformer encoders in our proposed MMTF-DES method. Despite the uncertainty in the predicted class label for individual multimodal transformer models, ViLT and VAuLT, our proposed method overcomes this limitation by using an effective early fusion strategy. 

\section{Conclusion and future work}
\label{sec:conclusionAndFutureDirection}
In this research, we have proposed a unified multimodal transformer-based architecture for the multimodal human desire understanding task where we fused two multimodal transformer models, ViLT and VAuLT. Using pairwise learning of the image-text pair setting of those multimodal transformer models, we have exploited the contextual relationship between title-caption pairs and the visual-contextual relationship between image-context pairs. That improves the intra-and-inter relationship of each image-text pair data in the human desire understanding task. We have also added multi-sample dropout layers on top of the concatenation-based early fused unified architecture of multimodal transformer models, improving the generalizability and speeding up training in the human desire understanding task. Experimental results on the benchmark dataset show that our proposed method outperforms all state-of-the-art methods. In this research, we propose a new task to better understand the human desire for image-text pairs data. We also discussed our system from various aspects, including modality dominance of image and text, early fusion impact, multi-sample dropout impact, and error analysis to validate the effectiveness of our proposed method. Fusing multimodal transformer-based architecture with a multi-sample dropout strategy on top of diverse multimodal transformer models has contributed to more robust representations of the image-text pair input while ensuring robustness throughout generating new representations. It is crucial for the human desire understanding task to improve performances by encouraging the model to extract more robust features. 

We plan to explore two other techniques. We intend to implement a bi-attention-based neural network, including both image and text-based attention, for better performance and greater efficiency in capturing the image and context. The second technique is domain-adaptive pre-training on the multimodal transformer models, where we need to feed relevant image-text pairs into the pre-trained multimodal transformer model. It may help to learn different human desire information effectively and improve the model performance.

\bibliographystyle{elsarticle-harv}
\bibliography{reference}

\end{document}